\documentclass[runningheads]{llncs}
\usepackage{graphicx}
\usepackage{array} 
\usepackage{subfigure} 
\usepackage{multirow} 
\usepackage{amsmath} 
 \usepackage[misc]{ifsym} 

\begin{document}
%
\title{Which and Where to Focus: A Simple yet Accurate Framework for Arbitrary-Shaped Nearby Text Detection in Scene Images\thanks{Supported by the Open Research Project of the State Key Laboratory of Media Convergence and Communication, Communication University of China, China (No. SKLMCC2020KF004), the Beijing Municipal Science \& Technology Commission (Z191100007119002), the Key Research Program of Frontier Sciences, CAS, Grant NO ZDBS-LY-7024, the National Natural Science Foundation of China (No. 62006221).}}
\titlerunning{A Simple yet Accurate Arbitrary-Shaped Nearby Text Detection Framework}
%
\author{Youhui Guo\inst{1,2} \and
Yu Zhou\inst{1,2}\textsuperscript{(\Letter)}   \and
Xugong Qin\inst{1,2} \and
Weiping Wang\inst{1}}
\authorrunning{Y. Guo et al.}
%
\institute{Institute of Information Engineering, Chinese Academy of Sciences, Beijing, China \and
School of Cyber Security, University of Chinese Academy of Sciences, Beijing, China \\
\email{\{guoyouhui, zhouyu, qinxugong, wangweiping\}@iie.ac.cn}}

\maketitle              
\begin{abstract}
Scene text detection has drawn the close attention of researchers. Though many methods have been proposed for horizontal and oriented texts, previous methods may not perform well when dealing with arbitrary-shaped texts such as curved texts. In particular, confusion problem arises in the case of nearby text instances. In this paper, we propose a simple yet effective method for accurate arbitrary-shaped nearby scene text detection. Firstly, a One-to-Many Training Scheme (OMTS) is designed to eliminate confusion and enable the proposals to learn more appropriate groundtruths in the case of nearby text instances. Secondly, we propose a Proposal Feature Attention Module (PFAM) to exploit more effective features for each proposal, which can better adapt to arbitrary-shaped text instances. Finally, we propose a baseline that is based on Faster R-CNN and outputs the curve representation directly. Equipped with PFAM and OMTS, the detector can achieve state-of-the-art or competitive performance on several challenging benchmarks.

\end{abstract}
\section{Introduction}
Scene text detection is a fundamental and crucial task in computer vision because it is an important step in many practical applications, such as scene text spotting, image/video understanding, and text visual question answering. 
Although immense progress has been made in previous excellent works, there still exist challenges in scene text detection due to large variations of shape, color, font, orientation, and scale, as well as the complex context of scene texts.

Recently, many segmentation-based approaches have been proposed for arbitrary shaped scene text detection, which performs text/non-text classification on each pixel and groups pixels into different instances. However, the borders between text regions and non-text regions are ambiguous and segmentation results are easily affected by nearby text. The approaches based on two-stage object detection framework  \cite{RD-IOD} with a more appropriate receptive field are still prevalent, in which the Region Proposal Network (RPN) is utilized to generate horizontal rectangle proposals and another detection head is designed to refine the proposals. In reality, text instances are arbitrary-shaped in scene images. The features extracted from horizontal rectangle proposals contain more background noise or even other text instances which is harmful to the classification and regression of the detection head. So, where should we focus on the feature map of each proposal?

Another existing issue is target confusion with the normal training scheme.
In the training process, the generated proposals are matched to the groundtruths by calculating the Intersection over Union (IoU). Under the widely used one-to-one matching rule, one proposal is forced to match the groundtruth with the highest IoU. However, the matched groundtruth with the hand-crafted rule may not be the best one. Moreover, when there exist multiple groundtruths with close IoUs to one proposal, i.e., the nearby text case, the network may get confused during learning, leading to suboptimal performance in inference. This is more likely to happen when text instances are close together which is common in scene images. So, which text instance should we choose to focus on?

\begin{figure}[]
\centering 
\includegraphics[width=0.65\textwidth]{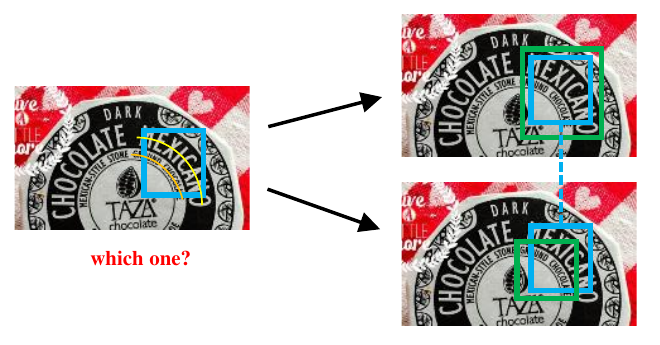}
\vspace{-0.3cm}
\caption{One proposal is assigned with multiple groundtruths in a one-to-many training scheme. Different curves represent different text instances. The blue and green boxes denote the proposals and groundtruths respectively.} 
\label{OMTS-example}
\vspace{-0.3cm}
\end{figure}

In this paper, we propose a simple and effective method to alleviate the aforementioned two problems for detecting arbitrary-shaped nearby text instances. For each proposal generated by RPN, we propose a Proposal Feature Attention Module (PFAM) to extract more useful features for further classification and regression. To alleviate the problem of target confusion during training, we propose a One-to-Many Training Scheme (OMTS) that one proposal can predict multiple text instances with no additional computation during inference. The illustration of the One-to-Many Training Scheme is shown in Fig. \ref{OMTS-example}.  
To verify the effectiveness of the proposed component, we integrate them into a baseline model which is based on Faster RCNN with curve regression for text detection. The experimental results show the effectiveness and robustness of our method on different datasets. 

The contributions of this work are summarized as follows:
\begin{itemize}
\item A novel One-to-Many Training Scheme is designed to alleviate the dilemma that the proposal may be confused about which instance to regress and no extra computations are involved in inference. 

\item We propose a Proposal Feature Attention Module for each proposal which can effectively extract more appropriate features for classification and regression.

\item 
A two-stage text detection framework with direct curve regression is proposed, and with the equipment of PFAM and OMTS, it achieves state-of-the-art or competitive performance on several benchmarks. Concretely, the proposed method achieves the F-Measure of 85.5\% on CTW1500, 87.7\% on Total-Text, 88.0\% on ICDAR2015, and 87.3\% on MSRA-TD500. Particularly, the proposed method outperforms the baseline with a large margin in the case of the nearby text.
\end{itemize}

\section{Related Work}

With the revival of neural networks, the majority of recent scene text detectors are based on deep neural networks. Generally, these methods can be roughly classified into three categories: regression-based, segmentation-based, and component-based methodologies.

\textbf{Regression-based approaches} are mainly inherited from some popular general object detection frameworks, which directly regress the bounding boxes of the text instances. TextBoxes++ \cite{DBLP:journals/tip/LiaoSB18} utilizes quadrilateral regression to detect multi-oriented text. Differently, EAST \cite{DBLP:conf/cvpr/ZhouYWWZHL17} directly detects the quadrangles of words in a pixel-level manner. FC$^2$RN \cite{FC2RN} uses a corner refinement network to produce a refined corner prediction. However, most of them show limited representation for irregular shapes, such as curved shapes.

\textbf{Segmentation-based approaches} are mainly inspired by semantic segmentation methods, which regard all the pixels within text bounding boxes as positive regions. 
Mask TextSpotter \cite{MaskTextSpotter} is the first end-to-end trainable arbitrary-shaped scene text spotter with a detection module based on Mask R-CNN. Qin et al.~\cite{WeaklySupervised} reduce the requirement of pixel-level annotations with weakly-supervised learning. Chen et al.~\cite{DomainAdaptive} propose a self-training framework with unannotated videos based on Mask R-CNN. 
Xiao et al.~\cite{SDM} propose a novel sequential deformation method to effectively model the line-shape of scene text. Performances of these methods are strongly affected by the quality of segmentation results.

\textbf{Component-based approaches} first detect individual text parts, then group them into texts with post-processing steps. CTPN \cite{CTPN} detects a sequence of fine-scale text proposals and connects those text proposals with RNN. CRAFT \cite{CRAFT} detects the text instances by exploring each character and affinity between characters. DRRG ~\cite{DBLP:conf/cvpr/ZhangZHLYWY20} constructs each text instance by a series of ordered rectangular components and utilize graph convolution network to reason the relations of those components. However, most of these methods need character-level annotations and depend on a good character detection network \cite{Constrained_Relation}. 

Different from most previous for arbitrary-shaped text, the proposed method directly produces a curve representation that requires only instance-level annotations.

\begin{figure}[t]
\centering 
\includegraphics[width=0.8\textwidth]{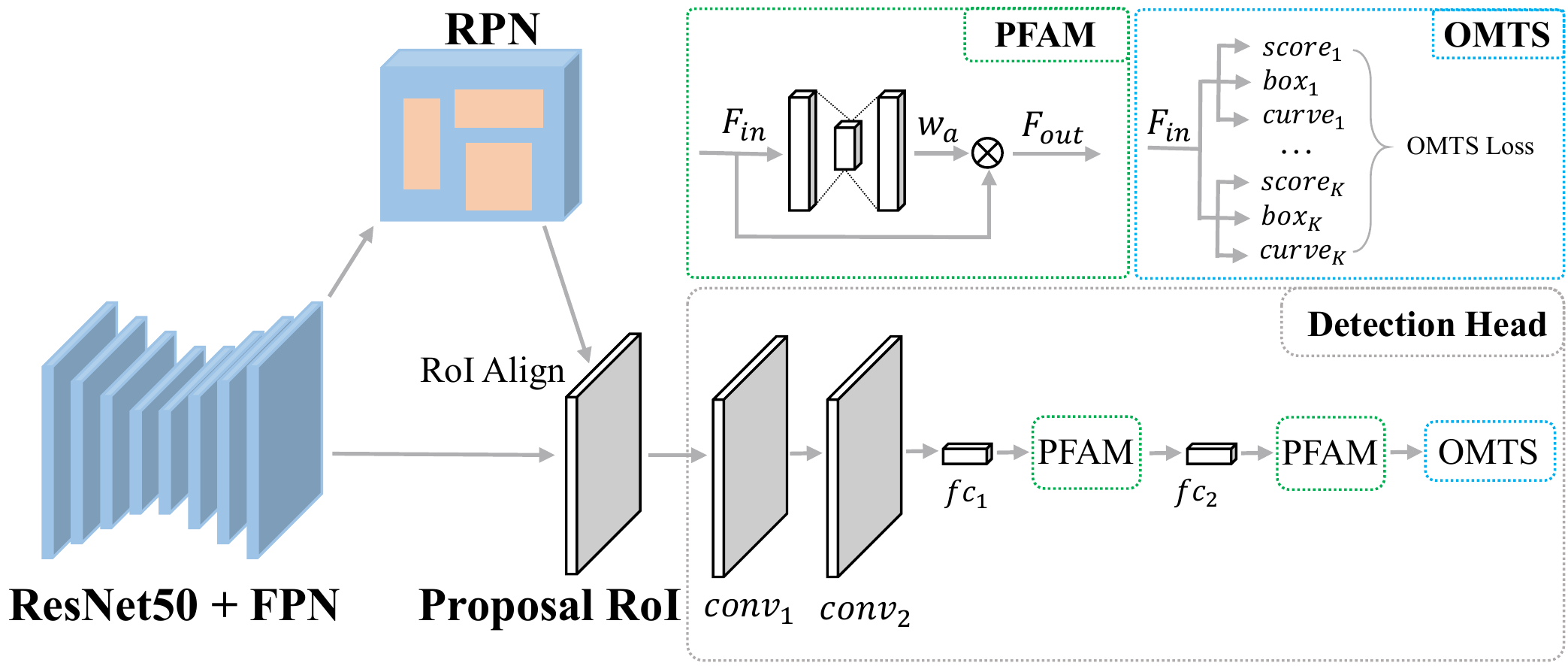}
\caption{The overall architecture. It consists of the Resnet50 backbone equipped with FPN, and the Detection Head working for each proposal.} 
\label{architecture}
\vspace{-0.3cm}
\end{figure}

\section{Proposed Method}
\subsection{Overview}
The overall architecture of our proposed method is shown in Fig. \ref{architecture}. The whole framework is based on Faster R-CNN. Firstly, the input image is fed into a feature pyramid backbone and a lot of proposals are produced by Region Proposal Network.
Then, the proposals are fed into Detection Head, in which a proposal feature attention module is designed to extract more appropriate features adaptively and a one-to-many training scheme predicts the multiple text instances with these features. The whole architecture is very simple and brings negligible computations. Next, we will elaborate the designed modules.

\subsection{One-to-Many Training Scheme}
In the training process of the second stage, the generated proposals are matched to the groundtruths by calculating the Intersection over Union (IoU). Under the widely used one-to-one matching rule, one proposal is forced to match the groundtruth with the highest IoU. However, the matched groundtruth with the hand-crafted rule may not be the best one. Moreover, when there exist multiple groundtruths with close IoUs to one proposal, i.e., the nearby text case, the network may get confused during learning, leading to suboptimal performance.
To alleviate this, we design a novel one-to-many training scheme which is inspired by CrowdDet \cite{DBLP:conf/cvpr/ChuZ0020}. Different from CrowdDet which aims at detecting highly-overlapped instances and recalling more targets in crowded scenes, the proposed OMTS is a training scheme with no additional computation during inference. With OMTS, one proposal can match with multiple groundtruths, which makes the learning process more efficient. When one proposal matches with only one groundtruth, the other targets are regarded as background. Specifically, for each proposal $p_{i}$, we match it with multiple groundtruth text instances $G(p_{i})$:
\begin{equation}
G(p_{i}) = \left \{ g_{j} \in \mathcal{G}|IoU(g_{j},p_{i})>\theta \right \},
\end{equation}
where $\mathcal{G}$ is the set of groundtruth instances and $\theta$ is a fixed threshold of IoU.

For each proposal $p_{i}$, we match it with $K$ groundtruth text instances by the fixed threshold $\theta$. Accordingly, we only need to produce $K$ sets of predictions $P(p_{i})$ at the end of the model, as shown below:
\begin{equation}
P(p_{i}) = \left \{ (\textbf{c}^{1}_{i},\textbf{r}^{1}_{i},\textbf{b}^{1}_{i}),(\textbf{c}^{2}_{i},\textbf{r}^{2}_{i},\textbf{b}^{2}_{i}),...,(\textbf{c}^{K}_{i},\textbf{r}^{K}_{i},\textbf{b}^{K}_{i}) \right \},
\end{equation}
where $\textbf{c}_{i}, \textbf{r}_{i}$ and $\textbf{b}_{i}$ denote predicted class confidence, the regression objective for axis-aligned rectangle box, and curve respectively. 

So we need to minimize the gap between predictions $P(p_{i})$ and groundtruth instances $G(p_{i})$ for each proposal $p_{i}$, the final $loss$ of OMTS defined as follows:

\begin{equation}
\begin{split}
\label{emdloss}
\mathcal{L}(p_{i}) = & \mathop{min}\limits_{\pi \in \Pi} \sum_{k=1}^{K} \left[ \mathcal{L}_{cls}(\textbf{c}^{(k)}_{i},g_{\pi_{k}}) + \mathcal{L}_{reg_{r}}(\textbf{r}^{(k)}_{i},g_{\pi_{k}}) \right. \\
 & \left. + \mathcal{L}_{reg_{b}}(\textbf{b}^{(k)}_{i},g_{\pi_{k}}) \right],
\end{split}
\end{equation}
where $\pi$ represents a certain permutation of $(1, 2, . . . ,K)$ whose $k$-th item is $\pi_{k}$, $g_{\pi_{k}} \in G_{p_{i}}$ is the $\pi_{k}$-th groundtruth text instance, $\mathcal{L}_{cls}, \mathcal{L}_{reg_{r}}$ and $\mathcal{L}_{reg_{b}}$ denote classification loss, box regression loss and curve regression loss respectively. Intuitively, the formulation in Eq. \eqref{emdloss} implies exploring all possible one-to-one matches between predictions and groundtruths, thus finding the “best match” with the smallest loss. This is a full permutation problem from 1 to K and its complexity is $O(K!)$, but considering the distribution of text instances in scene images, we make K=2 and this brings only a small computational complexity.

It is worth mentioning that when the proposal matches with a groundtruth and a background, we force the first output to predict the groundtruth and the second output to predict the background. By doing this, we only need to select the predictions of the first branch as the final result during the test, without adding any additional calculations.



\subsection{Proposal Feature Attention Module}
Region Proposal Network plays an important role in most two-stage object detection methods, which aims to generate horizontal proposals by coarse classification and regression based on the default anchors. 
However, text instances in natural images are arbitrary-shaped, the rectangles are not optimal for text instances because they contain a lot of background noise or even other text instances. Many methods obtain arbitrary-shaped text detection results by regression or segmentation in Detection Head, but very few methods focus on promoting the features of proposals.

To better adapt the features to arbitrary-shaped text instances, we propose a plug-and-play PFAM that can extract more appropriate features adaptively. Specifically, as shown in Fig. \ref{architecture}, with the feature generated by the previous fully connected (fc) layer denoted as $F_{in}$, we use a multi-layer perceptron (MLP) with one hidden layer and a sigmoid layer to produce the attention map $w_{a}$. With $F_{in}$ and $w_{a}$, the output of PFAM $F_{out}$ can be extracted by element-wise multiplication. The overall attention process can
be summarized as:
\begin{equation}
F_{out} = F_{in} \otimes \sigma(W_{MLP}(F_{in})),
\end{equation}
where $\sigma$, $\otimes$ and $W_{MLP}$ denote the sigmoid function, element-wise multiplication and the MLP weights. And the hidden dimension of the MLP is set to 64. 

Unlike other attention modules \cite{SENet,CBAM}, they focus on the global features of the entire image, while we pay more attention to the local features in each proposal.

\subsection{Representation for Arbitrary-Shaped Text Detection}
We first experiment with two kinds of representations for arbitrary-shaped text. One is a polygon with a fixed number of points and the other is the parametric Bezier curves. Specifically, we use 10 points to represent the long sides of the text instances (all 20 points) for the former; two Bezier curves \cite{ABCNet} are used to represent the long sides for the latter. The Bezier curve represents a parametric curve $B(t)$ defined as follows:
\begin{equation}
B(t)= \sum_{i=0}^{n} {\binom{n}{i}} P_{i}(1-t)^{n-i}t^{i},0\leq t\leq 1, \label{bezier}
\end{equation}
where $n$ represents the degree, ${\binom{n}{i}}$ is a binomial coefficient, and $P_{i}$ represents the $i-th$ control points. 
Practically, a cubic Bezier curve is sufficient to fit different kinds of arbitrary-shaped scene text for most of the existing datasets, so we use $n=3$ (all 8 points).

Based on Faster R-CNN, an extra branch is added to the fast R-CNN branch for polygon parameter regression. On CTW1500, they get the same performance with an F-measure of 83.7\%, which means comparable representation ability under the two presentations.  
The representation of the Bezier curves is more flexible with fewer parameters, so we use it in the rest part of the paper.

\subsection{Optimization}

\noindent \textbf{Learning targets}
Bezier-shaped text detection can be transformed to a regression with eight control points, which is consistent with the original bounding box regression in Faster R-CNN, so we adopt the parameterizations of the coordinates of eight control points as follows:
\begin{equation}
\begin{split}
t_{x} = (x_{e}-x_{p})/ w_{p}, t_{y} = (y_{e}-y_{p})/ h_{p}, \\
t^{*}_{x} = (x^{*}-x_{p})/ w_{p}, t^{*}_{y} = (y^{*}-y_{p})/ h_{p},
\end{split}
\end{equation}
where $x$, $y$, $w$, and $h$ denote the x-coordinate, x-coordinate, width, and height. Variables $x_{e}, x_{p},$ and $x^{*}$ are for the predicted control points, the proposals, and the groundtruths respectively (likewise for y). We regress the control points from the center of proposals. 

\noindent \textbf{Loss function}
The loss function $\mathcal{L}$ is formulated
as:
\begin{equation}
\mathcal{L} = \mathcal{L}_{rpn} + \mathcal{L}_{OMTS},
\end{equation}
where $\mathcal{L}_{rpn}$ is standard loss in RPN, $\mathcal{L}_{OMTS}$ is formulated as Eq. \eqref{emdloss}.


\section{Experiments}
We evaluate our approach on four standard datasets: CTW1500, Total-Text, ICDAR2015, MSRA-TD500, and compare with other state-of-the-art methods.

\subsection{Datasets}
The datasets used for the experiments in this paper are
briefly introduced below:

\noindent \textbf{Curve synthetic dataset} is a synthetic arbitrary-shaped scene text dataset proposed in \cite{ABCNet}, including 94,723 images containing a majority of straight text and 54,327 images containing mostly curved text. We use the annotations of this dataset to produce curve ground truth for pre-training our model.   \\ 
\noindent \textbf{CTW1500 dataset}  is another curved dataset that includes English and Chinese texts. CTW1500 contains 1,000 training images and 500 testing images. \\
\noindent \textbf{Total-Text dataset} is a dataset that includes horizontal, multi-oriented, and curved text. The annotations are bounding polygons. It consists of 1,255 training images and 300 testing images.\\
\noindent \textbf{ICDAR2015 dataset} contains 1000 images for training and 500 images for testing. The images are captured by Google Glass and the text accidentally appears in the scene. \\
\noindent \textbf{MSRA-TD500 dataset} is an oriented dataset that includes English and Chinese text instances with a large aspect ratio in natural scenes. It contains 300 training images and 200 testing images. Following the previous methods \cite{DB,ATRR,TextSnake}, we include extra 400 training images from HUST-TR400 \cite{DBLP:journals/tip/YaoBL14}.


\subsection{Implementation Details}

The data augmentation for the training data includes: (1) random crop, which we make sure that the crop size is larger than half of the original size and without any text being cut; (2) random scale training, with the short size randomly being chosen from (640, 672, 704, 736, 768, 800, 832, 864, 896) and the long size being less than 1600. In particular, we adopt random rotation with the range of $(-15^{\circ},15^{\circ})$ after random crop for ICDAR2015.

We use ResNet-50 with a Feature Pyramid Network which is initialized by ImageNet pre-trained model. Our model is pretrained on the Curve Synthetic dataset with 20 epochs. We use SGD as optimizer with batch size 1, momentum 0.9, and weight decay 0.0001 in training. We adopt warm-up in the initial 500 iterations. The initial learning rate is set to 0.0025 for all experiments.
In the inference period, we keep the aspect ratio of the test images and resize the test images by setting a suitable short side length for each dataset. The inference speed is tested with a batch size of 1 on a single GeForce RTX-2080Ti GPU.

\subsection{Ablation Study}
To verify the effectiveness of our proposed method, we conduct a series of comparative experiments on the ICDAR2015, CTW1500, and Total-Text. We first evaluate the newly proposed PFAM and OMTS in our method. Results are shown in Table 1. If not specified, all models are not pretrained for simplicity.
\begin{table}[!b]
\begin{center}
\caption{Effectiveness of PFAM and OMTS. 
“P”, “R”, and “F” refer to precision, recall and F-measure respectively.
}
\begin{tabular}{|c|l|c|c|c|}
\hline
Dataset & Method                  & P    & R    & F    \\ \hline
\multirow{6}{*}{Total-Text} & Baseline    & 87.8   & 83.0  & 85.3    \\ 
 & Baseline + PFAM(1fc)       & 88.3   & 83.3   & 85.7  \\ 
 & Baseline + PFAM(2fc)       & 87.4   & 84.1   & 85.7  \\ 
 & Baseline + OMTS  & 88.1   & 83.0  & 85.5  \\ 
 & Baseline + PFAM(1fc) + OMTS  & \textbf{88.9}    & 82.3   & 85.5   \\ 
 & Baseline + PFAM(2fc) + OMTS  & 87.5    & \textbf{84.7}   & \textbf{86.1}   \\ \hline
\multirow{6}{*}{CTW1500} & Baseline   & 85.6    & 81.8  & 83.7   \\ 
 & Baseline + PFAM(1fc)      & 85.4   & 82.5   & 83.9        \\ 
 & Baseline + PFAM(2fc)      & \textbf{86.6}   & 82.2   & 84.4        \\ 

 & Baseline + OMTS   & 86.1  & 82.6  & 84.3  \\ 
 & Baseline + PFAM(1fc) + OMTS  & 85.7    & 81.9   & 83.8   \\
 & Baseline + PFAM(2fc) + OMTS  & 86.2    & \textbf{83.6}   & \textbf{84.9}   \\ \hline

\end{tabular}
\end{center}
\vspace{-0.8cm}

\label{ablation}
\end{table}

\noindent \textbf{Baseline} 
We adopt the Bezier curve to represent text instances and integrate it into Faster R-CNN to obtain a very strong baseline. For a fair comparison, the baseline also adds two convolution layers. It achieves an F-measure of 85.3\% and 83.7\% on the Total-Text and CTW1500 respectively. 

\noindent \textbf{PFAM} 
First, we verify the effect of the number of PFAM on the model. PFAM(1fc) means that we only add a PFAM after the last fc layer and PFAM(2fc) means that we add a PFAM after each of the two fc layers. 
As shown in Table 1, Baseline with PFAM(1fc) and PFAM(2fc) can get a similar result on CTW1500 and Total-Text. But when we combine PFAM and OMTS, PFAM(2fc) can get a better result than PFAM(1fc). We believe that when using OMTS for training, the model needs a more appropriate feature, especially when there are nearby text instances. So we use PFAM(2fc) as the default setting. 

Concretely, compared to baseline, PFAM obtains 0.4\% and 0.7\% improvement in F-measure on TotalText and CTW1500 respectively. And compared to baseline with OMTS, PFAM can also obtain 0.6\% and 0.6\% improvement in F-measure on TotalText and CTW1500 respectively. 
For each axis-aligned box proposal, with a more appropriate feature extracted by PFAM for arbitrary-shaped text instances, we can obtain more accurate detection results.

\begin{table}[!t]
\begin{center}
\caption{The effectiveness of OMTS on the rotated datasets. $\dagger$ and $\ast$ denote CTW1500 and ICDAR2015 respectively.} 
\resizebox{0.7\textwidth}{!}{
\begin{tabular}{|c|c|c|c|c|c|c|c|c|c|c|}
\hline
\multirow{2}{*}{Dataset} & \multirow{2}{*}{OMTS} & \multicolumn{3}{c|}{Angle: 30} & \multicolumn{3}{c|}{Angle: 45} & \multicolumn{3}{c|}{Angle: 60} \\ \cline{3-11} 
   &                     & P           & R           & F           & P           & R           & F           & P           & R           & F           \\ \hline
\multirow{2}{*}{CTW1500} & $\times$              & 78.1        & 76.3        & 77.2        & 75.7        & 75.0        & 75.4        & 76.8        & 75.9        & 76.3        \\ 
 & $\surd$                 & \textbf{81.4}        & \textbf{77.4}        & \textbf{79.2}        & \textbf{78.3}        & \textbf{76.2}        & \textbf{77.2}        & \textbf{79.9}        & \textbf{76.2}        & \textbf{78.0}        \\ \hline
\multirow{2}{*}{ICDAR2015} & $\times$          & 85.9        & 81.2        & 83.5        & 85.2        & 79.6        & 82.3        & 85.0        & 81.7        & 83.3        \\ 
 & $\surd$              & \textbf{86.8}        & \textbf{81.7}        & \textbf{84.2}        & \textbf{85.8}        & \textbf{81.6}        & \textbf{83.7}        & \textbf{85.2}        & \textbf{82.4}        & \textbf{83.7}        \\ \hline
\end{tabular}
}
\end{center}
\label{OMTS}
\vspace{-0.5cm}
\end{table}

\noindent \textbf{OMTS} 
First, we evaluate our proposed OMTS on TotalText and CTW1500. As shown in Table 1, Compared to baseline, OMTS obtains 0.2\% and 0.6\% improvement in F-measure. When equipped with PFAM, OMTS obtains 0.8\% and 1.2\% improvement in F-measure on TotalText and CTW1500 respectively, which verifies the effectiveness of OMTS and the compatibility between OMTS and PFAM. 

To further verify the effectiveness of OMTS, We select an arbitrary-shaped and a multi-oriented text dataset respectively. So we augment the CTW1500 and ICDAR2015 test datasets with some specific angles, including $30^{\circ}$, $45^{\circ}$, and $60^{\circ}$. There are more situations in which the proposals are in a dilemma when the model tests on the rotated datasets. Specifically, the model is pretrained on the synthetic dataset and trained on the standard datasets with random rotation in [0,90]. 
The comparison of the performance with and without OMTS on the rotated datasets is shown in Table 2. The performance with OMTS consistently outperforms the one without OMTS by a large margin which shows the effectiveness of OMTS in more challenging scenarios. 

\begin{table}[]
\begin{center}
\caption{Detection results on the CTW1500 and Total-Text dataset. ${\dagger}$ means that the method pretrains on MLT2017.} 
\resizebox{0.6\textwidth}{!}{
\begin{tabular}{|c|c|c|c|c|c|c|c|c|c|c|c|}
\hline
\multirow{2}{*}{Method} & \multicolumn{4}{c|}{CTW1500} & \multicolumn{4}{c|}{Total-Text}  \\ \cline{2-9} 
& P           & R           & F  & FPS          & P           & R  & F & FPS \\ \hline
TextSnake \cite{TextSnake} &  67.9 & 85.3 & 75.6 & 1.1 & 82.7 & 74.5 & 78.4 & 1.1  \\ \hline
PSENet \cite{PSENet} & 84.8 & 79.7 & 82.2 & 3.9 & 84.0 & 78.0 & 80.9 &  3.9 \\ \hline
LOMO \cite{LOMO} & \textbf{89.2} & 69.6 & 78.4 & - & 88.6 & 75.7 & 81.6 & -  \\ \hline
CRAFT \cite{CRAFT} &  86.0 & 81.1 & 83.5 & - & 87.6 & 79.9 & 83.6 & - \\ \hline
PAN \cite{PAN} & 84.6 & 77.7 & 81.0 & 39.8 & 88.0 & 79.4 & 83.5 & 39.6 \\ \hline
Wang et al.~\cite{ATRR} & 80.1 & 80.2 & 80.1 & - & 80.9 & 76.2 & 78.5 &  \\ \hline
DB \cite{DB} & 86.9 & 80.2 & 83.4 & 22 & 87.1 & 82.5 & 84.7 & 32 \\ \hline
ContourNet \cite{ContourNet} & 83.7 & \textbf{84.1} & 83.9 & 4.5 & 86.9 & 83.9 & 85.4 & 3.8  \\ \hline
Zhang et al.~\cite{DBLP:conf/cvpr/ZhangZHLYWY20} & 85.9 & 83.0 & 84.5 & - & 86.5 & \textbf{84.9} & 85.7 & -  \\ \hline
Xiao et al.~${\dagger}$ \cite{SDM} & 85.8 & 82.3 & 84.0 & - & 89.2 & 84.7 & 86.9 & -  \\ \hline\hline
Ours & 88.8 &  82.4 & \textbf{85.5} & 18.7 & \textbf{90.7} & \textbf{84.9} & \textbf{87.7} &  10.6 \\ \hline

\end{tabular}
}
\end{center}
\label{table-ctw1500}
\vspace{-0.4cm}
\end{table}

\subsection{Comparison with Previous Methods}
\begin{figure}[t]
\centering 
\includegraphics[width=0.8\textwidth]{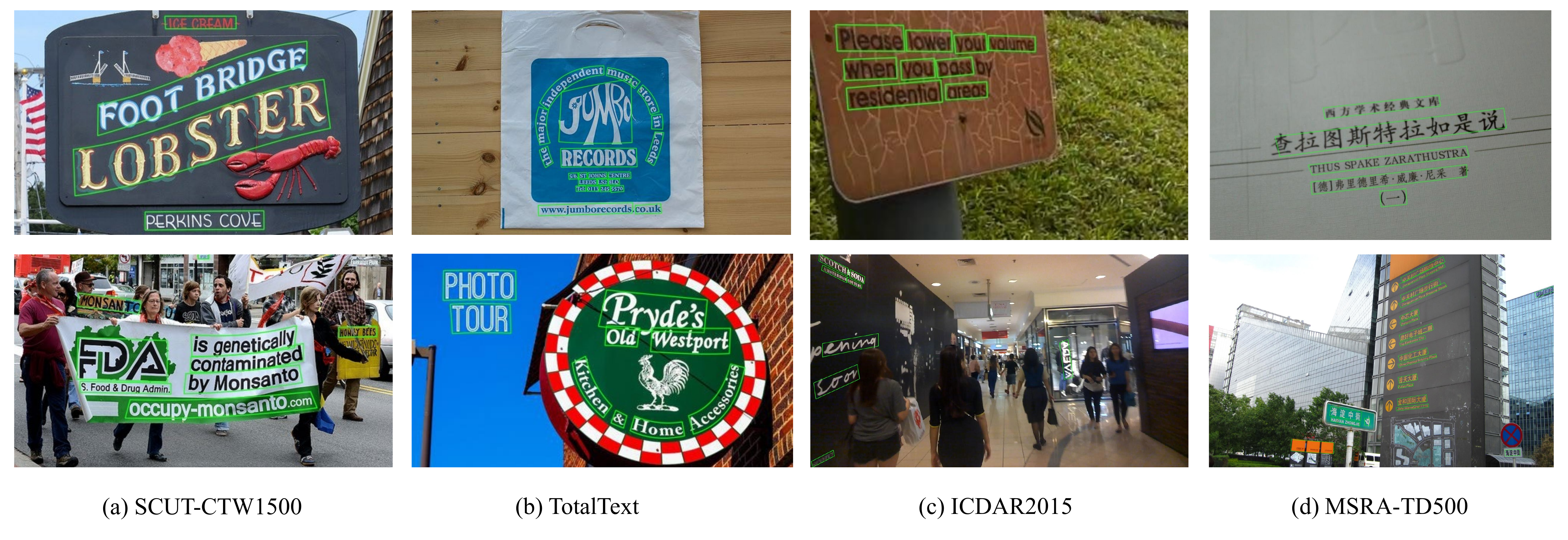}
\vspace{-0.4cm}
\caption{Some visualization results on different scene text datasets.} 
\label{visualization}
\vspace{-0.4cm}
\end{figure}

We compared our proposed method with previous methods on four datasets, including two datasets for curved texts, two datasets for multi-oriented texts. Some visualization results are shown in Fig. \ref{visualization}.

\noindent \textbf{Evaluation on CTW1500}
For CTW1500, we keep the short side of input images no less than 800 and the long side no more than 1333, meanwhile maintaining the aspect ratio. The comparison with the previous method is given in Table 3. Our proposed method achieves the state-of-the-art result of 88.8\%, 82.4\%, and 85.5\% in precision, recall, and F-Measure. Meanwhile, it also achieves a competitive speed (18.7 FPS). 

\noindent \textbf{Evaluation on Total-Text}
For Total-Text, We limit the short side of input images to no less than 1200 and the long side to no more than 2000 with the original aspect ratio. Also, our proposed method achieves a new state-of-the-art result of 90.7\%, 84.9\%, and 87.7\% in precision, recall, and F-Measure. The specific comparison is shown in Table 3. With the simple design of the entire framework, the speed is also achieved at 10.6 FPS. 

\noindent \textbf{Evaluation on ICDAR2015}
ICDAR2015 dataset is a multi-oriented text dataset that contains small and low-resolution text instances. Considering the small text instances, we double the width and height of the input image. As shown in Table 4, our proposed method achieves 88.0\% in F-Measure, which is higher than the most recent methods. It is worth noting that Xiao et al.~\cite{SDM} pretrain their model on MLT2017 and we pretrain our model on synthetic datasets. 

\begin{table}[]
\begin{center}
\caption{Detection results on the ICDAR2015 and MSRA-TD500 dataset. ${\dagger}$ means that the method pretrains on MLT2017.} 
\resizebox{0.6\textwidth}{!}{
\begin{tabular}{|c|c|c|c|c|c|c|c|c|c|}
\hline
\multirow{2}{*}{Method} & \multicolumn{4}{c|}{ICDAR2015} & \multicolumn{4}{c|}{MSRA-TD500}  \\ \cline{2-9} 
& P           & R           & F  &FPS          & P           & R  & F  &FPS\\ \hline
EAST \cite{DBLP:conf/cvpr/ZhouYWWZHL17} &  83.6 & 73.5 & 78.2 & 13.2 & 87.3 & 67.4 & 76.1 & - \\ \hline
TextSnake \cite{TextSnake} &  84.9 & 80.4 & 82.6 & 1.1 &  83.2 & 73.9 & 78.3 & 1.1 \\ \hline
CRAFT \cite{CRAFT} &   89.8 & 84.3 & 86.9 & -  & 88.2 & 78.2 & 82.9 & - \\ \hline
DB \cite{DB} & \textbf{91.8} & 83.2 & 87.3 & 12 &  91.5  & 79.2 & 84.9 & 32 \\ \hline
ContourNet \cite{ContourNet} & 86.1 & 87.6 & 86.9 & 3.5 & - & - & - & - \\ \hline
Zhang et al.~\cite{DBLP:conf/cvpr/ZhangZHLYWY20} & 84.7 & \textbf{88.5} & 86.6 & - & 88.1 & 82.3 & 85.1 & - \\ \hline
Xiao et al.~${\dagger}$ \cite{SDM} & 88.7 & 88.4 & \textbf{88.6} & - & - & - & - & - \\ \hline\hline

Ours & 90.6 & 85.4 & 88.0 & 6.5 & \textbf{92.7} & \textbf{82.5} & \textbf{87.3} &  19.0 \\ \hline

\end{tabular}
}
\end{center}
\label{table-ic15}
\vspace{-0.6cm}
\end{table}

\noindent \textbf{Evaluation on MSRA-TD500} 
To verify the robustness of detecting text instances with a large aspect ratio, we conduct experiments on the MSRA-TD500 dataset. In testing, we keep the short side of input images no less than 720 and the long side no more than 1200, meanwhile maintaining the aspect ratio. Detailed comparison results are shown in Table 4. Our proposed method achieves state-of-the-art performance with 92.7\%, 82.5\%, and 87.3\% in precision, recall, and F-Measure respectively, which outperforms other methods by a large margin and demonstrates that our proposed method is general and can detect oriented text instances with large aspect ratio in natural scenes. And our method is faster than most previous methods with 19.0 FPS. 

\section{Conclusion}
In this paper, we propose a simple yet accurate two-stage framework for detecting arbitrary-shaped scene nearby text instances. Based on Faster R-CNN, only minor modifications are made to the Detection Head for arbitrary-shaped text detection. We propose a proposal feature attention module and a one-to-many training scheme for more accurate detection, especially under the nearby text case. PFAM can extract more appropriate features effectively for each proposal. With the appropriate features, OMTS enables proposals to learn more appropriate text instances with negligible training time and no extra time during inference. Experiments on four benchmarks show that our proposed method has a good capability to detect horizontal, oriented, and curved scene texts. In the future, we will introduce the unsupervised methods \cite{VCP,Progressive_Cluster,VPRP} to text detection. Due to the detection result is friendly to the text recognition \cite{Gaussian,SEED}, we prefer to extend our proposed method to an end-to-end text spotting framework.

%
%
%
\bibliographystyle{splncs04}
\bibliography{icann}
%




\end{document}